\documentclass[5p,times,procedia]{elsarticle}
\usepackage{ecrc}

\volume{00}

\firstpage{1}

\journalname{Procedia CIRP}

\runauth{H. Wang et al.}

\jid{trpro}

\usepackage{amssymb}

\usepackage[figuresright]{rotating}

\usepackage[bookmarks=false]{hyperref}
    \hypersetup{colorlinks,
      linkcolor=blue,
      citecolor=blue,
      urlcolor=blue}

\usepackage{color,soul}

\usepackage{balance}

\begin{document}
\begin{frontmatter}



  \dochead{56th CIRP Conference on Manufacturing Systems, CIRP CMS '23, South Africa}%

  \title{Overview of Computer Vision Techniques in Robotized Wire Harness Assembly: Current State and Future Opportunities}


  \author[a]{Hao Wang*}
  \author[a]{Omkar Salunkhe}
  \author[b]{Walter Quadrini}
  \author[c]{Dan Lämkull}
  \author[d]{Fredrik Ore}
  \author[a]{Björn Johansson}
  \author[a]{Johan Stahre}

  \address[a]{Department of Industrial and Materials Science, Chalmers University of Technology, Hörsalsvägen 7A, SE-412 96 Gothenburg, Sweden}
  \address[b]{Department of Management, Economics and Industrial Engineering, Politecnico di Milano, piazza Leonardo da Vinci 32, 20133 Milan, Italy}
  \address[c]{Department of Manufacturing Technology, Volvo Car Corporation, SE-418 78 Gothenburg, Sweden}
  \address[d]{Department of Global Industrial Development, Scania CV AB, SE-151 87 Södertälje, Sweden}

  \aucores{* Corresponding author. Tel.: +46-(0)31-772-1202. {\it E-mail address:} haowang@chalmers.se}

  \begin{abstract}
    Wire harnesses are essential hardware for electronic systems in modern automotive vehicles.
    With a shift in the automotive industry towards electrification and autonomous driving, more and more automotive electronics are responsible for energy transmission and safety-critical functions such as maneuvering, driver assistance, and safety system.
    This paradigm shift places more demand on automotive wire harnesses from the safety perspective and stresses the greater importance of high-quality wire harness assembly in vehicles.
    However, most of the current operations of wire harness assembly are still performed manually by skilled workers, and some of the manual processes are problematic in terms of quality control and ergonomics.
    There is also a persistent demand in the industry to increase competitiveness and gain market share.
    Hence, assuring assembly quality while improving ergonomics and optimizing labor costs is desired.
    Robotized assembly, accomplished by robots or in human-robot collaboration, is a key enabler for fulfilling the increasingly demanding quality and safety as it enables more replicable, transparent, and comprehensible processes than completely manual operations.
    However, robotized assembly of wire harnesses is challenging in practical environments due to the flexibility of the deformable objects, though many preliminary automation solutions have been proposed under simplified industrial configurations.
    Previous research efforts have proposed the use of computer vision technology to facilitate robotized automation of wire harness assembly, enabling the robots to better perceive and manipulate the flexible wire harness.
    This article presents an overview of computer vision technology proposed for robotized wire harness assembly and derives research gaps that require further study to facilitate a more practical robotized assembly of wire harnesses.
  \end{abstract}


\end{frontmatter}




\section{Introduction}
\label{sec:intro}

\begin{figure}[tb]
  \centering
  \includegraphics[width=\linewidth]{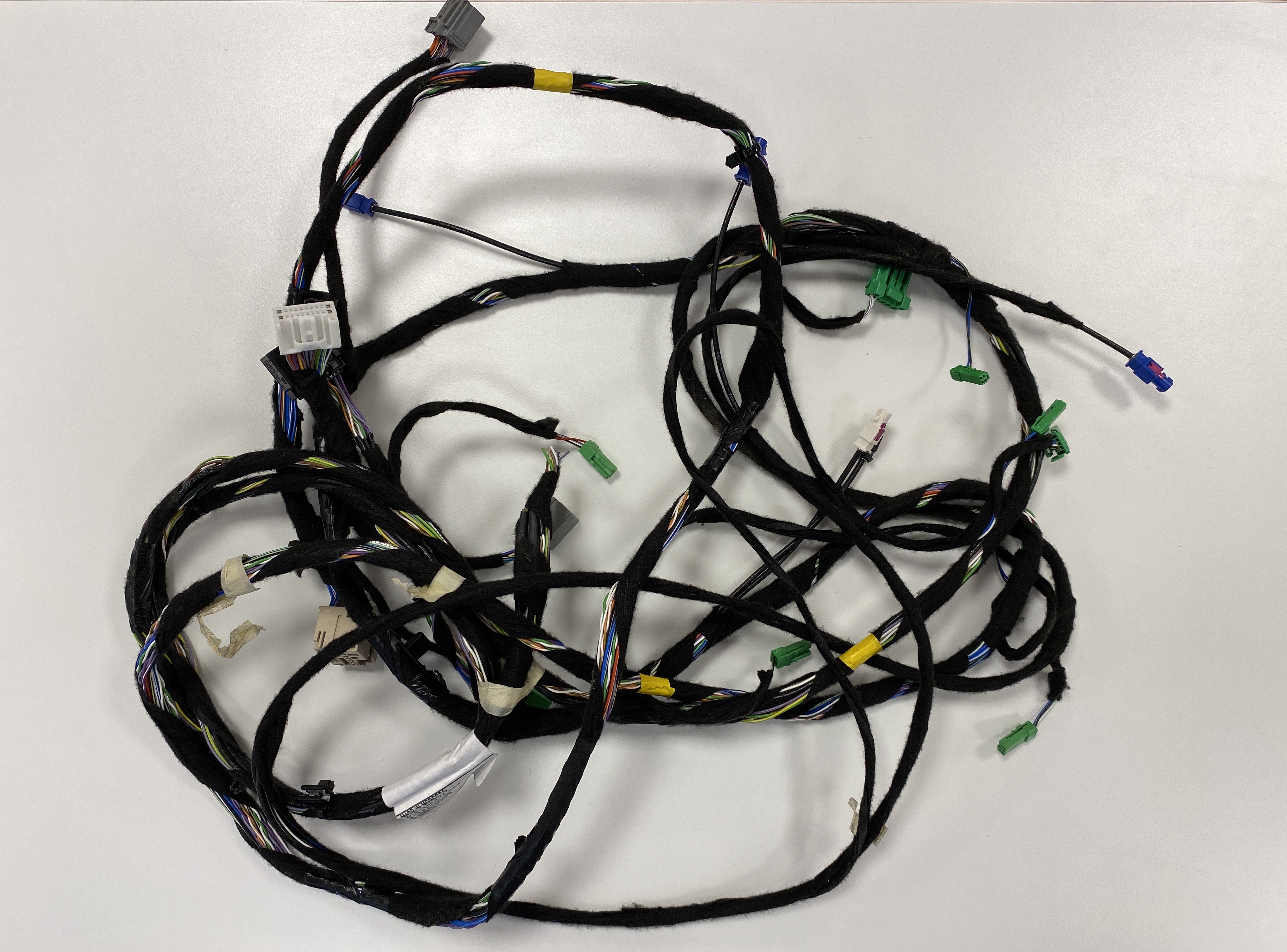}
  \caption{An example of an automotive wire harness.}
  \label{fig:wire_example}
\end{figure}

The shift towards electrification and autonomous driving in the automotive industry places increasing importance on the automotive electronic system, where more and more wire harnesses are installed as they are essential for connecting automotive electronics and supporting signal transmission in the electronic system.
This shift, in turn, makes the efficient, safe, and high-quality assembly of wire harnesses in vehicles critical for manufacturers to increase their competitiveness and gain more market share.
Fig.~\ref{fig:wire_example} illustrates an example of an automotive wire harness.

\begin{figure}[tb]
  \centering
  \includegraphics[width=\linewidth]{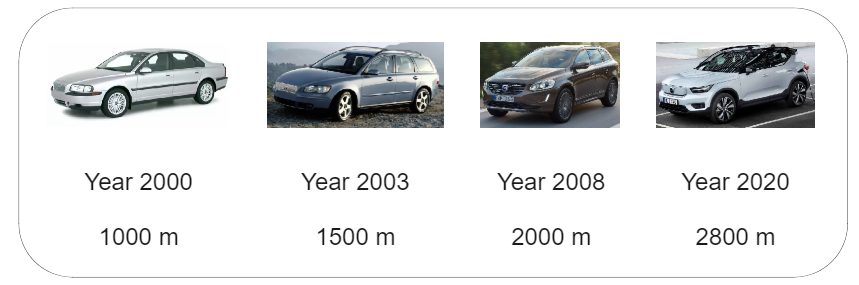}
  \caption{The heavily-increasing length of wire harnesses in passenger cars over time. (Courtesy of Volvo Car Corporation).}
  \label{fig:increasing_length}
\end{figure}

However, a large proportion of current automotive wire harness assembly operations remains manual and skill-demanding, leading to potential quality problems.
Some of the manual assembly processes also lead to severe ergonomic issues.
Meanwhile, the automotive industry persists in a consistent requirement for productivity.
Hence, it is desired to ensure assembly quality and improve ergonomics while keeping the utilization to optimal levels.

Robotized assembly has been implemented to facilitate automation in various industries.
Although different automation solutions have been proposed under simplified industrial configurations~\cite{sun2010robotic,jiang2010robotized}, robotized assembly of wire harnesses remains challenging in production due to the deformability of wire harnesses and short takt time.
In theory, as a special Deformable Linear Object (DLO) manipulation, robotized assembly of wire harnesses is challenging in modeling, state estimation, and operation~\cite{zhou2020practical,lv2022dynamic}.
The high flexibility of DLOs also limits transplanting methods for manipulating rigid objects~\cite{sanchez2018robotic,zhou2020practical} and makes it complex to design robot programs in production applications.
Other practical issues on general automation, such as risk management between robots and human operators, production management, and generalization of automation for multiple product variants on the same production line, also limit the robotization of wire harness assembly.

Computer vision presents a potential to the automotive industry and computer vision combined with robots can be the game changer in solving ergonomic issues whilst increasing quality and productivity.
Increasing research efforts in robotics have explored the complex DLO manipulation, addressing both theoretical research~\cite{saha2006motion,schulman2013tracking,tang2018framework} and engineering practices~\cite{di2009hybrid,jiang2010robotized,jiang2011robotized,jiang2015robotized,sanchez2018robotic,guo2022visual}.
Although various proposals under laboratory configurations have been discussed in previous studies~\cite{di2009hybrid,sun2010robotic,jiang2010robotized,jiang2011robotized,di2012vision,tamada2013high,song2017electric,zhou2020practical}, vision-based robotized wire harness assembly remains challenging in actual production, considering extracting image features from an intricate background~\cite{koo2008development} and fast recognition of wire harnesses for following robot operations~\cite{guo2022visual}.
Besides, vision-based automation solutions have not reached the robustness levels of human capabilities, especially under dynamic configurations in practical production.

This article aims at providing an overview of what type of vision systems have been proposed for robotized assembly of wire harnesses and exploring the future research directions toward a more practical vision system for robotized wire harness assembly.
Based on the previous literature, this study identifies three major trends for future research, including 1) the adaptation of more advanced object detection and recognition algorithms to facilitate the robotized wire harness assembly; 2) new product design on the components of wire harnesses to facilitate the visual machine perception; and 3) the evaluation of proposed vision systems in actual production to examine the practicality and reliability.

In the rest of this article, Section~\ref{sec:basics} introduces the basics of automotive wire harnesses and the task for automation of the final assembly of wire harnesses according to current assembly workflow in production, followed by the methodology of this study in Section~\ref{sec:method}.
Then, Section~\ref{sec:component} and Section~\ref{sec:structure} describe vision systems proposed in previous research for manipulating components of wire harnesses and perceiving the structure of wire harnesses, respectively, followed by discussions in Section~\ref{sec:discussion}.
Section~\ref{sec:concl} concludes the article with summarized findings and outlooks on future research direction.
The follow-up research of this study is briefly introduced in Section~\ref{sec:future}.

\section{Basics of automotive wire harnesses}
\label{sec:basics}

A wire harness has a tree-like structure, consisting of a bundle of routed cables with various components, such as clamps and connectors, which are used to transmit current or signals within electrical equipment~\cite{aguirre1994economic,tilindis2014effect}.
The usage of wire harnesses keeps enlarging in modern vehicles due to the increasing number of electronic devices installed for various functions, which can be reflected in the heavily-increasing length of automotive wire harnesses installed in automobiles over time, as illustrated in Fig.~\ref{fig:increasing_length}.

Wire harnesses can be categorized according to the location of installation, for example, engine harness, instrument panel harness, floor harness, and bumper harness.
Based on the observation in a car manufacturing plant, the current wire harness assembly in the passenger cabin of a passenger vehicle can be divided into (1) automated preprocessing on the packed wire harness; (2) transfer of the preprocessed wire harness into the car body using lifting equipment; (3) disentangle and route the wire harness manually; and (4) fix and connect the wire harness manually.
The various heavyweights of wire harnesses and several far-reaching assembly positions pose challenges to human operators to manipulate and assemble them manually in the final assembly in terms of quality and ergonomics.
In addition, several high-voltage wire harnesses, especially in an electric car, need to be handled more carefully regarding assembly quality, safety, and reliability, making automation desired and in urgent need.

Robots are widely implemented as automated solutions in manufacturing and have demonstrated tremendous potential for the automation of wire harness assembly.
Besides mechanical control, a robot needs to know the position of picking and placing the target objects before operation.
Considering the complexity of wire harness assembly, computer vision may facilitate a more flexible positioning and prove considerable aid in pick-n-place and assembly operations.

\section{Methodology}
\label{sec:method}

To acquire an overview of what vision-based technology has been discussed for robotized wire harness assembly, this study first implemented an inquiry on the Scopus database with the search string, \textit{TITLE-ABS-KEY((wir* OR cabl*) AND (harness* OR bundl*) AND assembl*)}.
According to the main purpose of this study, the searching results were examined thoroughly by the co-authors to select the articles focusing on the final assembly of wire harnesses onto other products and proposing vision systems for the robotized assembly.
The selected articles were then analyzed to identify the current state and future research needs.
Besides, only articles in English were included in the analysis, and secondary studies, i.e., review and conference review, were excluded from the analysis.

\section{Component manipulation}
\label{sec:component}

Some previous studies have focused on the manipulation of different components of wire harnesses, including clamps~\cite{koo2008development,jiang2010robotized,jiang2011robotized,jiang2015robotized} and connectors~\cite{di2009hybrid,sun2010robotic,di2012vision,tamada2013high,song2017electric,yumbla2020preliminary,zhou2020practical}, for achieving robotized assembly of wire harnesses, as listed in Table~\ref{tab:component}.

\begin{table*}[tb]
  \caption{Vision systems in articles for manipulation on components of wire harnesses.}
  \begin{tabular*}{\hsize}{@{\extracolsep{\fill}}lllll@{}}
    \toprule
    Component & Article & Type of cameras & Location of cameras & Number of cameras \\
    \colrule
    Clamp & \cite{koo2008development} & - & Hand-eye & 4 \\
    & \cite{jiang2010robotized,jiang2011robotized} & CCD cameras & Global-fixed + Hand-eye & 10 fixed + 6 on end-effectors \\
    & \cite{jiang2015robotized} & Point Grey Firefly MV & Hand-eye & 1 \\
    Connector & \cite{tamada2013high} & MC1362, Mikrotron & Global-fixed & 1 \\
    & \cite{yumbla2020preliminary} & RealSense D435, Intel & Hand-eye & 1 \\
    & \cite{zhou2020practical} & Industrial cameras & Global-fixed + Hand-eye & 1 fixed + 2 on robot arms \\
    & \cite{di2009hybrid} & In-Sight 5100 & Global-fixed & 1 \\
    & \cite{sun2010robotic} & CCD cameras & Global-fixed & 2 \\
    & \cite{di2012vision} & CCD cameras & Global-fixed & 2 \\
    & \cite{song2017electric} & FL2G-13S2C-C, PGR & Hand-eye & 1 \\
    \botrule
  \end{tabular*}
  \label{tab:component}
\end{table*}

\subsection{Clamp insertion}

Clamps are used for fixing wire harnesses on target locations.
Four articles discussed clamp insertion onto an automobile instrument panel frame with different vision systems~\cite{koo2008development,jiang2010robotized,jiang2011robotized,jiang2015robotized}, where CCD cameras were implemented to recognize clamps so that the end-effector on a robot arm could reach, grasp, and manipulate the detected clamps.
All four studies implemented hand-eye vision systems by mounting different numbers of cameras on the end-effectors of robot arms~\cite{koo2008development,jiang2010robotized,jiang2011robotized,jiang2015robotized}.
Two of them also adopted global vision systems with multiple cameras fixed around the operation area to support recognition, for example, avoiding occlusion~\cite{jiang2010robotized,jiang2011robotized}.

Koo et al.~\cite{koo2008development} first proposed to facilitate the clamp recognition and manipulation by installing cubic clamp covers with markers, whose poses were recognized by identifying the markers with SIFT~\cite{lowe1999object,lowe2004distinctive} using stereo vision systems consisting of two CCD cameras with different focal lengths mounted on the end-effectors of two robot arms.
Later, Jiang et al.~\cite{jiang2010robotized} and Jiang et al.~\cite{jiang2011robotized} improved the clamp covers to a cylinder-like shape with more markers from ARToolKit~\cite{artoolkit,kato1999marker} and implemented a global vision system comprising ten fixed cameras with different angles surrounding the work-frame alongside hand-eye cameras to relief the occlusion problem.
Furthermore, Jiang et al.~\cite{jiang2015robotized} proposed to replace visual clamp detection with a tracing operation.
Nonetheless, one wrist CMOS camera on the right robot arm remained to estimate the pose of the cover later following the similar design of pattern recognition in Jiang et al.~\cite{jiang2010robotized} and Jiang et al.~\cite{jiang2011robotized}.

\subsection{Connector mating}

Mating of connectors is critical in wire harness assembly to ensure quality and functionality, which is complicated with demanding manipulation accuracy and intricate structures and non-rigid materials of connectors~\cite{sun2010robotic}.
There are seven articles proposing various vision systems for different tasks of connector mating, including state estimation~\cite{tamada2013high,yumbla2020preliminary,zhou2020practical}, vision-guided mating~\cite{di2012vision,song2017electric}, and fault detection~\cite{di2009hybrid,sun2010robotic}.

State estimation of the 3D geometric information is essential for robot control in vision-based robotized assembly.
Tamada et al.~\cite{tamada2013high} implemented a global-fixed high-speed camera to recognize the types and poses of connectors and monitor the mating process by detecting the corners of connectors with image processing at high speed.
Zhou et al.~\cite{zhou2020practical} used three cameras, one fixed global camera and one hand-eye camera per arm of a dual-arm robot, to achieve a rough locating-then-fine positioning detection based on MobileNet-SSD~\cite{howard2017mobilenets} and CAD model registration.
Instead of using 2D cameras, Yumbla et al.~\cite{yumbla2020preliminary} adopted a RealSense D435 depth camera, where RGB images were first processed for connector detection with image processing methods and then the depth image was integrated to obtain the 3D geometry.

Dedicated to the vision-guided mating process, Di et al.~\cite{di2012vision} designed a scheme for monitoring relative motions between two connectors with two mutually perpendicular cameras based on basic pattern matching.
Song et al.~\cite{song2017electric} proposed a pattern matching-based visual servoing for locating connector headers using a hand-eye camera and markers on connector headers.

Fault detection has also been discussed to ensure the quality of grasp and insertion so that the control system could react to problematic manipulations earlier.
Di et al.~\cite{di2009hybrid} adopted an In-Sight 5100 camera to detect faults in grasp and insertion by checking the relative translational and rotational displacements between the gripper and the connector, which was further improved by Sun et al.~\cite{sun2010robotic} by adding one camera perpendicular to the first one to supplement the displacement detection.

\section{Structure perception}
\label{sec:structure}

Recently, three studies have discussed different tasks in perceiving the structure of wire harnesses, including interpretable classification of branches~\cite{kicki2021tell}, 3D profile extraction~\cite{nguyen2021novel}, and wire recognition~\cite{guo2022visual}, as listed in Table~\ref{tab:structure}.

\begin{table*}[tb]
  \caption{Vision systems in articles for perceiving the structure of a wire harness.}
  \begin{tabular*}{\hsize}{@{\extracolsep{\fill}}lllll@{}}
    \toprule
    Article & Purpose & Type of cameras & Location of cameras & Number of cameras \\
    \colrule
    \cite{kicki2021tell} & Interpretable classification & RealSense D435, Intel & Global-fixed & 1 \\
    \cite{nguyen2021novel} & 3D profile extraction & Helios Time-of-Flight camera & Hand-eye & 1 \\
    \cite{guo2022visual} & Visual recognition & RGB-D & - & - \\
    \botrule
  \end{tabular*}
  \label{tab:structure}
\end{table*}

Kicki et al.~\cite{kicki2021tell} focused on the interpretable classification of wire harness branches and proposed an RGB-D image dataset of four branches of an automotive wire harness captured by a global-fixed RealSense D435 depth camera.
This research compared different networks sharing the same Downsample layer from ERFNet~\cite{romera2018erfnet} for different data modalities and evaluated the impact of elastic transformation and pre-training with inpainting task.
Saliency maps based on class activation mapping (CAM)~\cite{zhou2016learning} were adopted to visualize the interpretability of the classification results.

Nguyen and Yoon~\cite{nguyen2021novel} proposed to obtain the 3D geometry of a wire harness through clamp detection and profile tracking and correction using a hand-eye depth camera (Helios Time-of-Flight Camera) mounted on the right UR3 robot to guide the selection of next robot picking point.

Guo et al.~\cite{guo2022visual} proposed an RGB-D-based segmentation and estimation for aircraft wire harness recognition, where the complete segmented wires were obtained through supervoxel over-segmentation, segmentation based on Cartesian distance, color similarity, and bending continuity, and estimation with Gaussian Mixture Model~\cite{reynolds2009gaussian} on the raw point cloud data acquired by an RGB-D camera.

\section{Discussion}
\label{sec:discussion}

In general, previous research efforts designed different vision systems to recognize different components of wire harnesses~\cite{koo2008development,di2009hybrid,jiang2010robotized,sun2010robotic,jiang2011robotized,di2012vision,tamada2013high,jiang2015robotized,song2017electric,yumbla2020preliminary,zhou2020practical}, estimate the state of sub-processes~\cite{di2009hybrid,di2012vision,song2017electric,kicki2021tell,guo2022visual}, and detect errors in assembly~\cite{di2009hybrid,sun2010robotic} so that the control system of robotized wire harness assembly could conduct manipulations on wire harnesses and monitor the operation process according to the real-time configurations.

Previously, the most focused components of wire harnesses were clamps~\cite{koo2008development,jiang2010robotized,jiang2011robotized,jiang2015robotized} and connectors~\cite{di2009hybrid,sun2010robotic,di2012vision,tamada2013high,yumbla2020preliminary,zhou2020practical,song2017electric}.
Earlier studies on clamp insertion took advantage of designed clamp covers with unique markers to facilitate detection and manipulation~\cite{koo2008development,jiang2010robotized,jiang2011robotized,jiang2015robotized}.
However, these studies~\cite{koo2008development,jiang2010robotized,jiang2011robotized,jiang2015robotized} focused on the assembly of wire harnesses to an automobile instrument panel, which could be significantly different from other assembly locations that demand a different number of wire harnesses, for example, in the engine room or cabin.
Clamp covers would occupy space in other installation areas with more compact installation, and post-assembly cover removal would lead to further challenges.
Thus, new product designs to facilitate the visual machine perception and robotic manipulation are desired.
Other studies on connectors discussed how to detect connectors~\cite{tamada2013high,yumbla2020preliminary,zhou2020practical} and monitor the assembly process~\cite{di2009hybrid,sun2010robotic,di2012vision,song2017electric}.
2D image recognition was primarily adopted in previous research except for one that introduced depth information alongside RGB color information captured by an RGB-D camera~\cite{yumbla2020preliminary}.
However, the 2D RGB image was processed initially to detect connectors, and the depth information of detected connectors was simply extracted from the depth image~\cite{yumbla2020preliminary}.

Considering more recent studies on perceiving the structure of wire harnesses with RGB-D and ToF cameras~\cite{kicki2021tell,nguyen2021novel,guo2022visual} and the increasingly advanced and affordable imaging technology, it is promising to implement depth or 3D cameras to acquire spatial information and conduct detection by processing 3D information directly.
The recent renaissance of convolutional neural networks (CNN) and the successful development of deep learning in computer vision research~\cite{lecun2015deep} also facilitate numerous research on learning-based 2D and 3D computer vision problems~\cite{guo2021deep,minaee2022image,zou2023object}, which inspires the adaptation of latest object detection and recognition algorithms to the future computer vision-driven robotized assembly of wire harnesses to facilitate the detection, manipulation, and tracking of various components of wire harnesses.

In addition, although previous studies have demonstrated the potential of vision-based solutions for robotized wire harness assembly tasks, the practicality and reliability of the proposals in practical applications remain unknown.
The actual production has a strict requirement on the production rate, which makes the operating speed and quality of the vision system critical and validation of vision systems in actual circumstances necessary.
Thus, future studies on the evaluation of computer vision techniques for robotized wire harness assembly in practical manufacturing configurations are desired and necessary to assure the practicality and reliability as well as the successful integration of the vision systems in actual manufacturing.

\section{Conclusion}
\label{sec:concl}

In conclusion, previous studies have noticed and explored various computer vision-based solutions for different tasks in robotized assembly of wire harnesses, including vision-guided manipulation of different components and visual machine perception of the structure of wire harnesses.
Yet, further development on robotized assembly and the enabling computer vision-based techniques are needed to promote a sound work environment, less ergonomic stress, and sustainable production, especially considering the increasing number of cables in high-tech products, e.g., cars, as depicted above.
The state of the art of computer vision technology in robotized wire harness assembly was reviewed in this paper and some future research opportunities were discussed, including:

\begin{itemize}
  \item Developing new learning-based computer vision algorithms to exploit 3D information captured by depth or 3D cameras to facilitate the detection, manipulation, and tracking in robotized wire harness assembly;
  \item Evaluating the practicality and reliability of vision systems in actual production to promote the integration of vision systems into practice;
  \item Exploring new product designs of wire harnesses to enable a more efficient component detection and manipulation without affixing additional parts.
\end{itemize}

\section{Future work}
\label{sec:future}

In the future, vision-guided robotized manipulations on various components of wire harnesses will be proposed and evaluated in both laboratory and practical scenarios.
A systematic literature review will also be conducted to better understand the state of the art of vision-guided robotized wire harness assembly and distinguish the challenges in the task.


\section*{Acknowledgements}

This work was supported by the Swedish innovation agency, Vinnova, and the strategic innovation program, Produktion2030, under grant number 2022-01279.
The work was carried out within Chalmers’ Area of Advance Production.
The support is gratefully acknowledged.

\bibliography{mybib}
\bibliographystyle{elsarticle-harv}


\end{document}